\documentclass{article}

\usepackage{times}
\usepackage{graphicx} 
\usepackage{subfigure} 

\usepackage{algorithm}
\usepackage{algorithmic}

\usepackage{hyperref}

\usepackage[accepted]{icml2016}

\usepackage{amsfonts}
\usepackage{amssymb}
\usepackage{amsmath}
\usepackage{amsthm}
\usepackage{latexsym}
\usepackage{graphicx}
\usepackage{verbatim} 

\usepackage{algorithmic}
\usepackage{algorithm}

\icmltitlerunning{Dynamic Task Allocation for Crowdsourcing Settings}

\begin{document} 
	
	\twocolumn[
	\icmltitle{Dynamic Task Allocation for Crowdsourcing Settings}
	
	\icmlauthor{Angela Zhou}{angelaz@princeton.edu}
	\icmladdress{Princeton University, Princeton, NJ 08544}
	\icmlauthor{Irineo Cabrerors}{cabreros@princeton.edu}
	\icmladdress{Princeton University, Princeton, NJ 08544}
	\icmlauthor{Karan Singh}{karans@princeton.edu}
	\icmladdress{Princeton University, Princeton, NJ 08544}
	
	\icmlkeywords{crowdsourcing, mutual information, active learning, task assignment }
	
	\vskip 0.3in
	]
	
	\begin{abstract} 
	We consider the problem of optimal budget allocation for crowdsourcing problems, allocating users to tasks to maximize our final confidence in the crowdsourced answers. Such an optimized worker assignment method allows us to ``boost" the efficacy of any popular crowdsourcing estimation algorithm. We consider a mutual information interpretation of the crowdsourcing problem, which leads to a stochastic subset selection problem with a submodular objective function. We present experimental simulation results which demonstrate the effectiveness of our dynamic task allocation method for achieving higher accuracy, possibly requiring fewer labels, as well as improving upon a previous method which is sensitive to the proportion of users to questions. 
	\end{abstract} 
	
	\section{Introduction}
	\label{motivation}
The rationale of crowdsourcing is to leverage the ``wisdom of the crowd" in soliciting some kind of response or suggestion. Crowdsourcing allows for the annotation of large research corpuses for training data-intensive models such as deep neural networks. Platforms such as Mechanical Turk allow for the systematic and programmatic implementation and assignment of crowdsourcing tasks. The crowdsourcing estimation problem is especially difficult because both the reliabilities of users and the true answers to the questions are unknown. 

\subsection{Previous Work} 
Dawid and Skene studied this estimation problem in 1979 in the context of responses to surveys [\citenum{Dawid79}]. In the Dawid-Skene stochastic model, each user has an associated reliability score dictating the probability with which the user answers a binary question correctly. The joint estimation of user reliabilities and correct answers in this model is very well studied in statistics, even for the case of $k$-ary questions. Popular approaches use Expectation Maximization to find the true labels maximizing the likelihood of estimated answer labels and reliabilities. Recent work uses spectral methods (based on eigenvalues of the assignment graph) to initialize the iterative expectation maximization algorithm, proving optimality convergence rates of such a scheme [\citenum{Zhang2014}]. Another approach achieving state-of-the-art performance finds the labeling via a minimax conditional entropy approach [\citenum{Zhou2014}].

Most analysis of the estimation algorithms assume random sampling. However, conceivably, intermediate estimates of the reliabilities of workers can inform better worker-task assignments which lead to higher quality final estimates. A 2011 paper by Karger, Oh, and Shah on budget-optimal crowdsourcing tracks multiple instances of distinct assignment graphs but assumes a highly specific "spammer-hammer" model where all users either answer randomly or correctly. Another approach models the assignment problem as a Markov Decision Process and derives the Optimistic Knowledge Gradient, which computes the conditional expectation of choosing certain workers, assuming a Beta-Bernoulli prior on the reliability of each worker [\citenum{Chen2013}]. However the assignment scheme requires extensive recomputation and updating between each individual question-worker assignment, an unrealistic frequency of model updating. 

\textbf{Classical Dawid-Skene Model} \label{DSmodel}
In the classical Dawid-Skene model, abbreviated ``D-S", there are $n$ users and $m$ questions with correct answers $\{-1, +1\}$. (Without loss of generality we may map the answers to $\{0, 1\}$.) Let $G\in \{0,1\}^{n \times m}$ denote a bipartite matrix indicating if the $i^{th}$ user answered the $j^{th}$ question. This is typically called the {\em assignment} matrix. Additionally, let $a\in \{-1,+1\}^m$ be a vector which captures the correct answer for each of the $m$ questions. Let $p\in [0,1]^n$ be a vector denoting each user's reliability, for $n$ users. Let $A\in\{-1,0,1\}^{n\times m}$ be a stochastic {\em answer} matrix such that
\[
A_{ij}=\left\{
\begin{array}{ll}
0  & \text{if }G_{ij}=0\\
a_j & \text{with probability } p_i, \text{if }G_{ij}=1\\
-a_j & \text{with probability } 1-p_i, \text{if }G_{ij}=1\\
\end{array}
\right.
\]
\section{Method for Quasi-Online Task Allocation}
We derive an improved task allocation scheme and model the crowdsourcing problem as an optimization problem with an information-theoretic objective, since we want to assign workers to tasks to gain the most information about the true label of the question.

\begin{figure}[h!]\centering
	\includegraphics[scale=0.4]{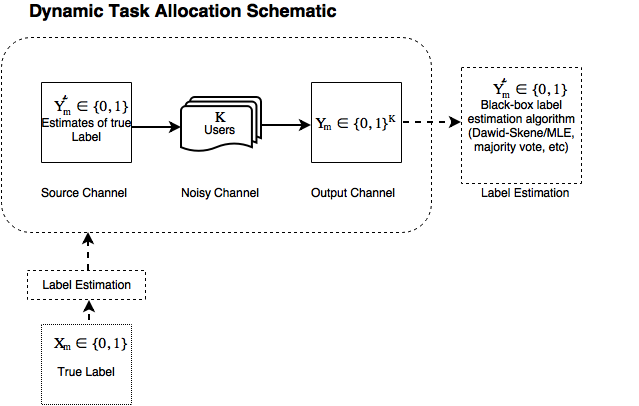}
	\caption[Schematic diagram of dynamic task allocation scheme.]{Schematic diagram of the dynamic task allocation scheme.} 
	\label{fig-dynamic-task-allocation}
\end{figure}

Previous work (unpublished) by the same authors \cite{Cabreros2015} proposes a two-step estimation method that uses a budget parameter, $s \in [0,1]$, in two stages. Running a crowdsourcing estimation algorithm when half of the budget has been allocated yields estimates of the true answers, which may be used with a mixture model on the topics of questions to estimate reliabilities of each user, $\tilde{\textbf{F}}$. Then sample users more likely to provide correct answers for specific questions during the remaining portion of the budget to arrive at a final $\textbf{A}$. The final answers matrix $\textbf{A}$ is then used as input for the same black-box estimation algorithm to arrive at the final estimates, as depicted in Figure  \ref{fig-dynamic-task-allocation}. 

How do we decide which questions require more budget allocated? We consider information-theoretic metrics: mutual information sums over all possible outcomes of the random variables X, Y, and measures the mutual dependence between two variables by quantifying the ``amount of information" obtained about one random variable by observing another \citep{Cover2006}. A variant of the mutual information between realized outcomes of the random variables $X=x, Y=y$ is the \textit{pointwise mutual information} which has found applications in statistical natural language processing. We define another variant of mutual information, \textbf{partial mutual information}, between a random variable on the source channel, $X$, and \textit{observed outcomes} of $Y = y$, where we only integrate over the randomness of the unknown source channel. \textbf{Partial Mutual Information}, pMI(X; Y), is defined between random variable $X$ and outcomes of random variable $Y = y$. 
	\[pMI(X; Y) = \sum_{x \in X} p(x,y) \log\left(\frac{p(x,y)}{p(x)p(y)}\right)\]

In our notation: 
\begin{align*}
& \mathbb{P}[X_m = 1, \{Y_m\}] = \prod_{i \vert y_i = 1} F_{im} \prod_{j \vert y_j = 0} (1 - F_{jm})\\
\end{align*}

However, solving the cardinality-constrained integer program to assign users to questions is most likely NP-hard, as \cite{Krause2012} reduce a similar formulation of an entropy minimization problem to independent set. The best thing to do in a `one-shot' setting if our remaining budget $s$ is less than the number of questions $m$ is to choose the $s$ best relative improvements in the partial mutual information, which we denote as $\Delta pMI_{rel}(v; X_m ;Y_m)$ for the improvement with respect to querying user $v$. Estimating $F_{ij}$ has already computed a ``nested optimization" and found which user is the best to assign to a certain question. 
\section{Results} 
\begin{figure}[h]
	\includegraphics[scale=0.55]{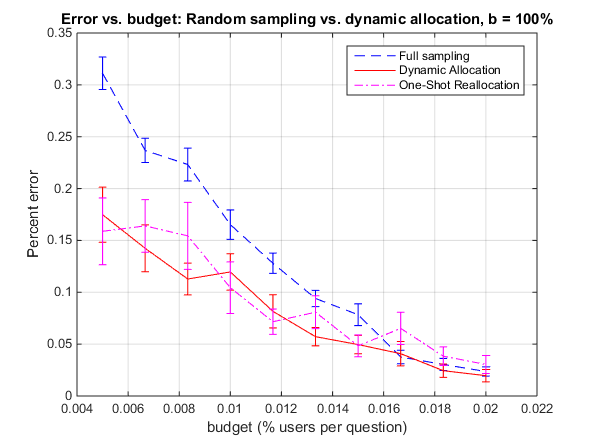}
	\caption{Comparison of error vs. budget percentages for the dynamic allocation method (blue) against fully random sampling and Dawid-Skene (dashed red) estimator. Note that for small budgets the dynamic allocation scheme significantly improves estimation error. For large budgets both methods are highly accurate. Error bars are 1.96 standard errors over 25 trials (confidence level of 95\%)}
	\label{fig-crowdsourcing-budget-v-error}
\end{figure}

We evaluate the error rates of three methods: (1) running Dawid-Skene estimation on a budget $s$ randomly sampled, and (2) running Dawid-Skene estimation on a budget $\frac{s}{2}$ randomly sampled and assigning the remaining budget via the one-shot allocation, and (3) method (2) but assigning budget via the dynamic task allocation. We evaluate the error rates from 20 random samples of questions and user reliabilities for 1000 users and 100 questions, assuming a mixture of 2 topics, over 10 different budgets from assigning $0.5\%$ of users to each question to $2\%$ coverage in Figure \ref{fig-crowdsourcing-budget-v-error}. 

\begin{figure}[h]		\includegraphics[scale=0.55]{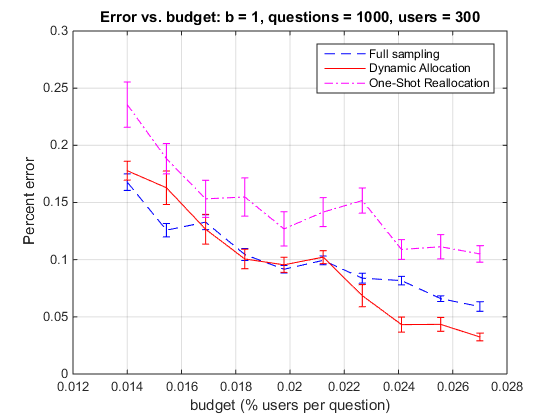}
	\caption{Comparing sampling policy performance when varying the number of questions, 10 sample trials each. Error bars are one standard deviation. Dynamic task allocation (red solid) is competitive with full sampling (blue dashed), while one-shot learning (magenta dashed) performs poorly as the number of questions increases} 
	\label{fig:crowdsourcing-varying-m}   
\end{figure}

We also consider how the dimensions of the estimation problem impact the performance of these sampling policies in Figure \ref{fig:crowdsourcing-varying-m}: when users outnumber questions, the one-shot allocation does poorly while the new dynamic user allocation is still robust. 

 However, examining the intermediate accuracies yielded by the dynamic task allocation method indicate that the strengths of this method lie in the ability to achieve better estimation accuracy performance with fewer samples, and terminate the estimation process early. Can we use the estimates of mutual information in an optimal stopping framework to develop more data-efficient crowdsourcing estimation algorithms? ``Efficiency" will be measured with regards to random sampling, where we want to show our method is $(1+\epsilon)$ efficient using $(1-\delta)$ samples as needed by random sampling, for any chosen $\delta$. We will consider this question in the full paper and conjecture that one point of connection between the channel capacity is with the Fisher Information. Analysis would proceed by considering the minimax convergence rates of the D-S estimator, proved in \cite{Gao2013}, to analyze the disadvantage of using fewer samples. 
\section{Conclusions} 
We model task assignment in the crowdsourcing problem and develop a probabilistic model for a ``partial mutual information" criterion which yields a one-step lookahead policy of which questions to ask next. We implement a batch estimation policy which exhausts the budget by querying an additional label for each question, re-estimating $\tilde{F}_{est}$, and using these updated estimates to re-compute the label estimates. We are able to show significant improvement in estimation accuracy for small budgets. Our dynamic task allocation scheme is also robust for estimation schemes with higher ratios of questions to users, where the one-shot learning policy suffers higher error than random sampling. The advantage of such a scheme is that the mutual information heuristic we develop may be extended to evaluate early termination of the estimation process. 

\bibliographystyle{icml2016}
\bibliography{refs}

\end{document}